%% file: main.tex
\documentclass[runningheads]{llncs}
\usepackage[T1]{fontenc}
\usepackage{subcaption}
\usepackage{float}
\usepackage{xcolor}

\usepackage{wrapfig}
\usepackage{mathtools}
\usepackage{amsmath}
\usepackage{amssymb}

\usepackage{graphicx} 
\usepackage{cancel}
\usepackage{float}
\input{macros}

\title{Exploring Flexible Scenario Generation in Godot Simulator}

\author{Daniel Peraltai\inst{1} \and
Xin Qin\inst{1}}
\institute{California State University, Long Beach,CA, USA
\email{Daniel.Peralta02@student.csulb.edu}, 
\email{xin.qin@csulb.edu}}

\begin{document}
\maketitle
\begin{abstract}

Cyber-physical systems (CPS) combine cyber and physical components engineered to make decisions and interact within dynamic environments. Ensuring the safety of CPS is of great importance, requiring extensive testing across diverse and complex scenarios. To generate as many testing scenarios as possible, previous efforts have focused on describing scenarios using formal languages to generate scenes.  In this paper, we introduce an alternative approach: reconstructing scenes inside the open-source game engine, Godot. We have developed a pipeline that enables the reconstruction of testing scenes directly from provided images of scenarios. These reconstructed scenes can then be deployed within simulated environments to assess a CPS. This approach offers a scalable and flexible solution for testing CPS in realistic environments. 

\end{abstract}

\keywords{Scenario Generation  \and Simulation \and Signal Temporal Logic}

\section{Introduction}


The verification of cyber-physical system (CPS) safety is critical to ensuring the reliability and resilience of autonomous technologies. CPS often operates in safety-critical domains such as transportation, healthcare, and industrial automation \cite{ivanov2020case}, \cite{renkhoff2024survey}, \cite{bak2022neural}, where rigorous and comprehensive verification methodologies are essential \cite{qin2024statistical}. In autonomous driving, scenario description and enumeration have proven effective in capturing diverse driving conditions and detecting potential system bugs within simulation environments \cite{scenic}, \cite{scenic3}. Scenic \cite{fremont2023scenic}, a widely used tool for scenario generation, facilitates the enumeration of varied driving scenarios, enabling extensive testing of autonomous systems under diverse conditions.

A notable limitation arises, however, when trying to enumerate roads and the placement of buildings within simulators. Platforms such as CARLA \cite{dosovitskiy2017carla} and AirSim \cite{airsim2017fsr}, developed using the Unreal Engine, lack native support for flexible map customization, particularly for road networks. Map creation requires manual construction and compilation by developers for each iteration. Unlike agents that spawn in a map, map components (e.g., roads) present inherent complexity and variability, making exhaustive enumeration nontrivial. Additionally, the initial construction process is not easily described by current formal languages. This limitation may reduce the variability of environments in which CPS can be tested, creating challenges for subsequent testing and validation. As a result, guarantees can only be made on certain maps, raising the question of how to generate additional maps and refine them to expose system weaknesses.

We present a scenario reconstruction pipeline capable of generating road models within a simulator from images. Our approach demonstrates that image-based techniques can roughly replicate original road geometries. In contrast to traditional methods that model the environment as a single 3D object, our technique generates roads sequentially, allowing real-time human interaction and on-the-fly adjustments within the simulator.


We further illustrate how this approach integrates with formal methods, enabling data regeneration and fine-tuning of reconstructed roads to meet specific accuracy thresholds. This method diverges from AI-generated approaches by utilizing formal methods to guide targeted adjustments, ensuring greater precision and control. A detailed comparison between our method and AI-based techniques is provided in Section \ref{sec:related-work}. 

In summary, the main contributions of this paper are as follows:
\begin{itemize}
\item Development and implementation of a pipeline that extracts environmental information from images and imports it into simulation environments, leveraging the latest advancements in game engines.
\item Propose integrating formal method specifications to quantify and guide the refinement of the generated environments.
\end{itemize}

\section{Preliminaries}
\subsection{Signal Temporal Logic} Signal Temporal Logic (STL) \cite{maler2004} is a widely adopted formalism used to express safety specifications across various CPS applications \cite{stl_survey}. STL defines the behavior of signals through logical formulas. These formulas are constructed over signal predicates of the form $f(\traj) \geq c$ or $f(\traj) \leq c$, where $\traj$ represents a signal, $f: \reals^n \rightarrow \reals$ is a real-valued function, and $c \in \reals$. The syntax of STL is provided in Eq.~\eqref{eq:stlsyntax}. We assume that the interval $I=[a,b]$ satisfies $a,b \in \reals^{\ge 0}$ and $a \leq b$, with $\sim \in \{\leq, \geq\}$.
\begin{equation}
\label{eq:stlsyntax}
\varphi,\psi:= \mathit{true} \mid
               f(\traj) \sim c \mid
               \neg \varphi \mid
               \varphi \wedge \psi \mid
               \varphi \vee \psi \mid
               \eventually_I \varphi \mid
               \always_I \varphi \mid
               \varphi~\until_I~\psi
\end{equation}
In the syntax described above, $\eventually$ (eventually), $\always$ (always), and $\until$ (until) are temporal operators. For any $t \in \nnreals$ and interval $I = [a,b]$, we denote $t + I$ as $[t+a, t+b]$. For a signal $\traj$ at a specific time $t$, the notation $(\traj, t) \models \varphi$ indicates that $\traj$ satisfies $\varphi$ at time $t$. We let $\traj \models \varphi$ to represent $(\traj, 0) \models \varphi$.

\subsection{Distance Metrics} 
We equip the set of splines $\mathcal{Y}$ with a function $d:\mathcal{Y}\times\mathcal{Y}\to \mathbb{R}$ that quantifies distance between the splines. A natural choice of $d$ is a signal metric that results in a metric space $(\mathcal{Y},d)$. We use general signal metrics such as the metric induced by the $L_p$ signal norm for $p\ge 1$. Particularly,  define
    \(d_p(y_1,y_2):=\Big(\int_{\mathbb{T}} \|y_1(t)-y_2(t)\|^p\mathrm{d}t\Big)^{1/p}\)
so that the $L_\infty$ norm can also be expressed as $d_\infty(y_1,y_2):=\sup_{t\in \mathbb{T}} \|y_1(t)-y_2(t)\|$.

\begin{wrapfigure}{r}{0.8\linewidth}
    \centering
        \vspace{-1cm}
    \begin{subfigure}{0.35\linewidth}
        \centering
        \includegraphics[width=\linewidth, trim={2cm, 1.64cm, 2.1cm, 1.2cm}, clip]{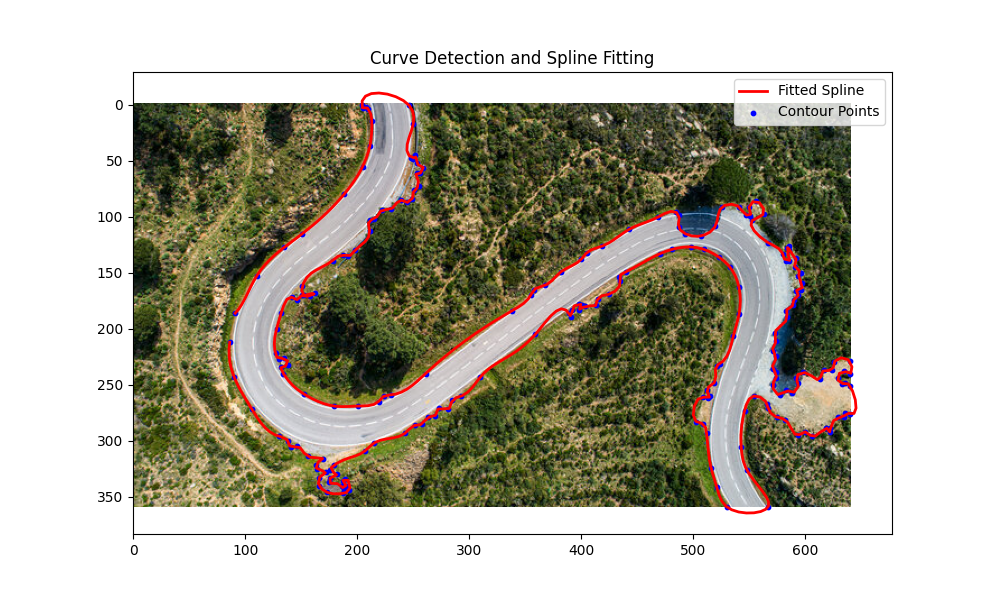}
        \caption{Fit splines}
        \label{fig-contour-shadow}
    \end{subfigure}
    \begin{subfigure}{0.325\linewidth}
        \centering
        \includegraphics[width=\linewidth, trim={0cm, 0cm, 0cm, 0cm}]{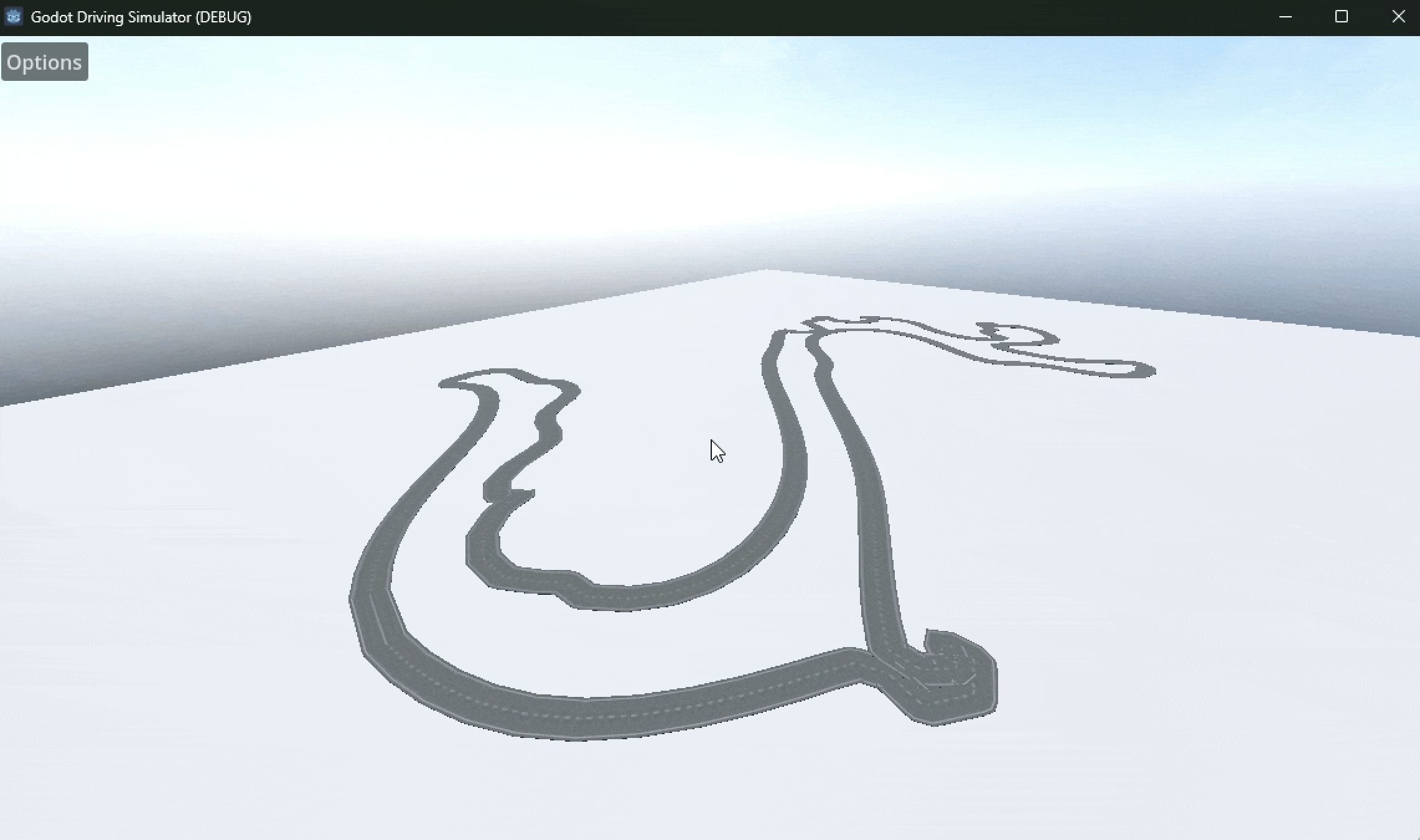}
        \caption{Generate in godot}
    \end{subfigure}
    \begin{subfigure}{0.3\linewidth}
        \centering
        \includegraphics[width=\linewidth, trim={0cm, 0cm, 7.9cm, 1cm}, clip]{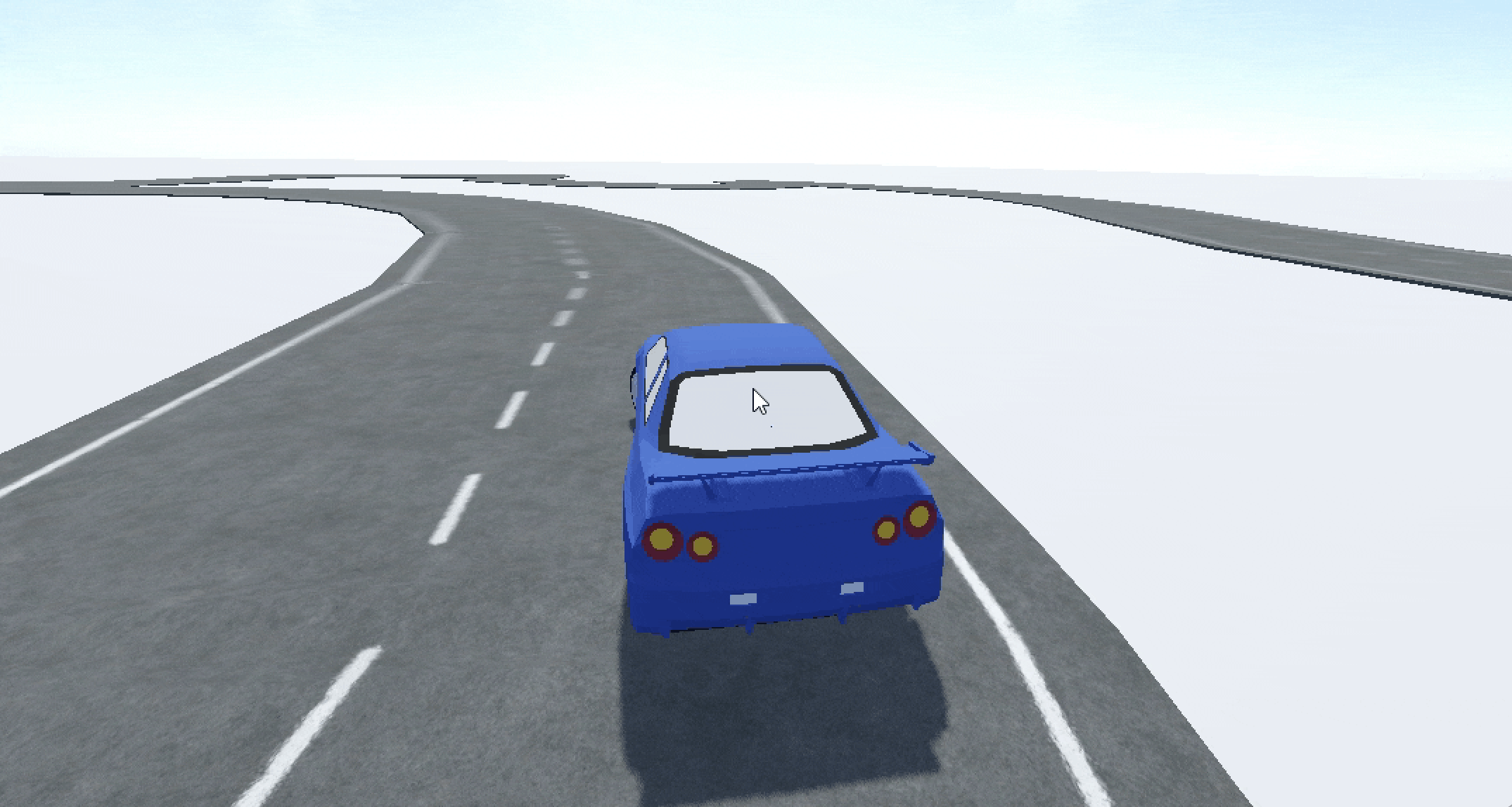}
        \caption{Put car on road}
    \end{subfigure}
    \label{fig:case-1}
        \vspace{-0.5cm}
    \caption{Extract the road from the image and recreate it Godot.}
    \vspace{-0.6cm}
\end{wrapfigure}

\section{Flexible Scenario Generation in Godot}
Godot allows for the generation of tiles and objects entirely through code. Given a list of coordinates, we can create the map purely programmatically.

\subsection{Extract the road}
The initial stage of the pipeline follows common practices, applying robust image processing and computer vision techniques—adjusting brightness, contrast, and sharpness, applying Gaussian blur, and performing contour extraction and spline fitting—to detect roads. 


\subsection{Modifying the Road}

We aim to generate variations of the road, but not all variations are feasible. Variability is introduced through sinusoidal perturbations of the spline. We use Signal Temporal Logic (STL) to constrain the modifications applied to the road, ensuring that the newly generated road remains meaningful for testing.

The following specifications define constraints on road perturbations and spline distances. Let $e_1$ represent the error added to the original spline, modeled as a random variable through a sampling-based method, and let $d_1$ denote the distance between generated splines.  

Specification $\varphi_1$ constrains road modifications to remain below a threshold of 10, written in $\varphi_1 := \always(e_1 < 10)$.
Specification $\varphi_2$ ensures that the distance between splines consistently stays below 10, written as $\varphi_2 := \always(d_1 < 10)$. Specification $\varphi_3$ specifies that if the spline error perturbation exceeds the threshold, the distance must eventually reduce to below 10, and the error must also decrease to under 10, with $\varphi_3 := \always\left( (e_1 > 10) \implies \eventually_{[t_1, t_2]} \always (d_1 < 10 \wedge e_1 < 10) \right)$. The time interval in STL can specify which segment of the spline should exhibit this behavior, allowing humans to write specifications that constrain the shape of the road. We can filter perturbations that meet design requirements by applying the monitoring algorithm of the STL formula.

\section{Experiments and Results}
We propose a pipeline consisting of two components. First, we analyze the input image to extract the desired information. Second, we pass this information to the simulator, which generates the corresponding environment. This approach allows us to fine-tune scenarios within the simulator with great flexibility. We can directly add controllers and various agents to enable interactions, and the design is adaptable for connecting multiple scenarios seamlessly in the future.

\paragraph{Scenario Reconstruction Using Image}

\begin{wrapfigure}{r}{0.6\linewidth}
    \centering
    \begin{subfigure}{0.9\linewidth}
        \centering
\includegraphics[width=\linewidth]{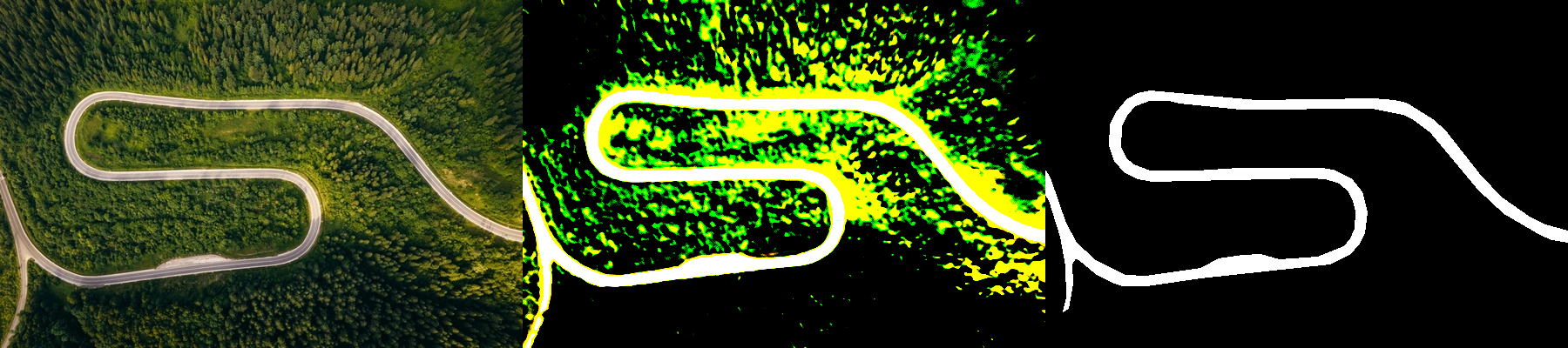}
 \caption{Extract the contour}
 \label{fig:extract-contour}
    \end{subfigure}
    \begin{subfigure}{0.48\linewidth}
        \centering
 \includegraphics[width=\linewidth, trim={2cm, 1.6cm, 2.1cm, 1.2cm}, clip]{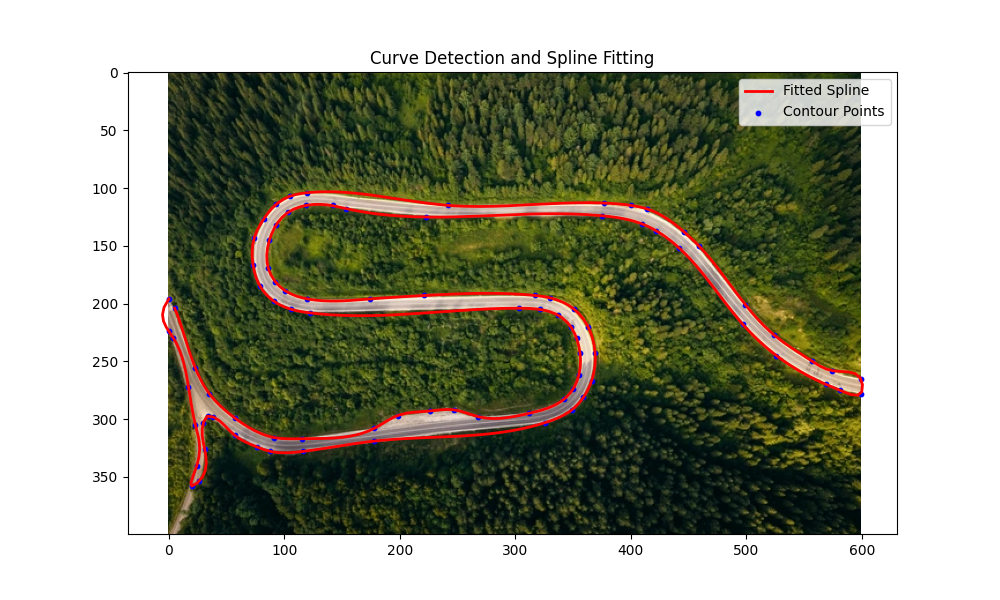}
        \caption{Fit splines}
    \end{subfigure}
        \begin{subfigure}{0.48\linewidth}
        \centering
 \includegraphics[width=\linewidth, trim={2cm, 1.6cm, 2.1cm, 1.2cm}, clip]{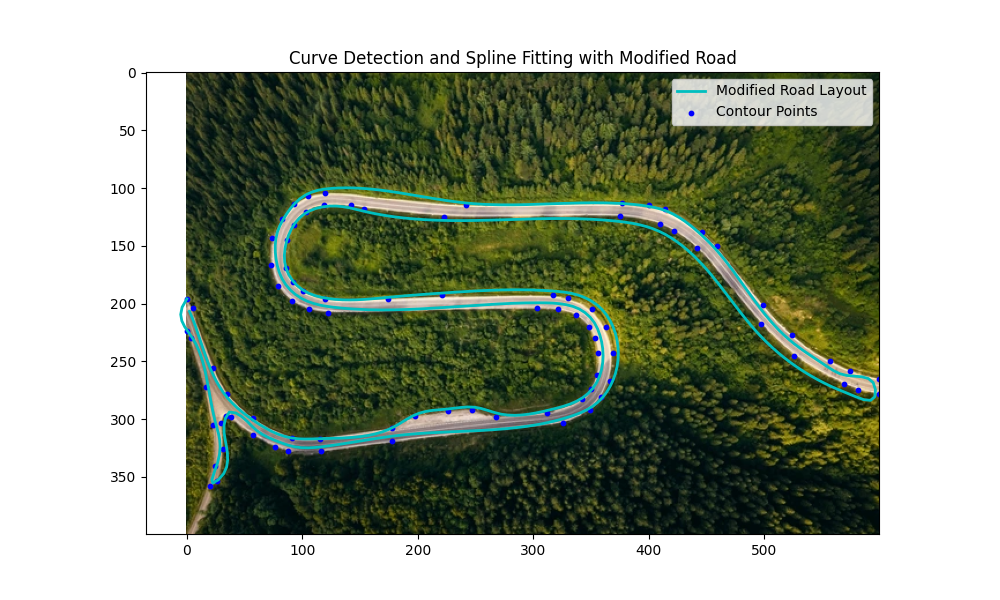}
        \caption{Modify the splines}
    \end{subfigure}
            \begin{subfigure}{0.45\linewidth}
        \centering
 \includegraphics[width=\linewidth]{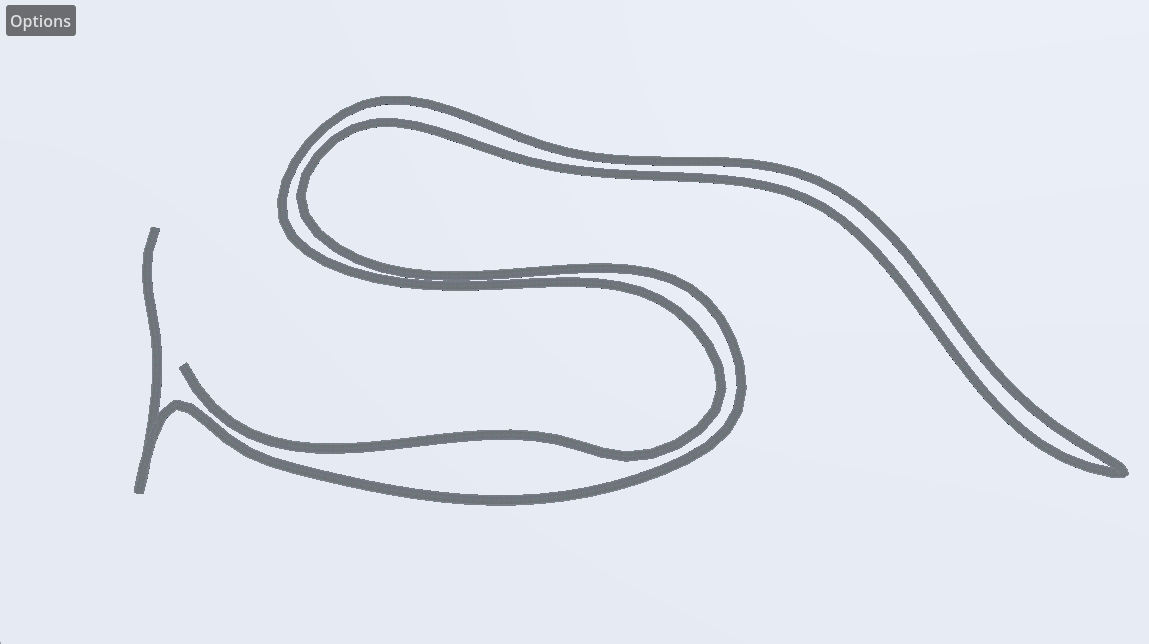}
        \caption{Modify the splines}
    \end{subfigure}
                \begin{subfigure}{0.45\linewidth}
        \centering
 \includegraphics[width=\linewidth]{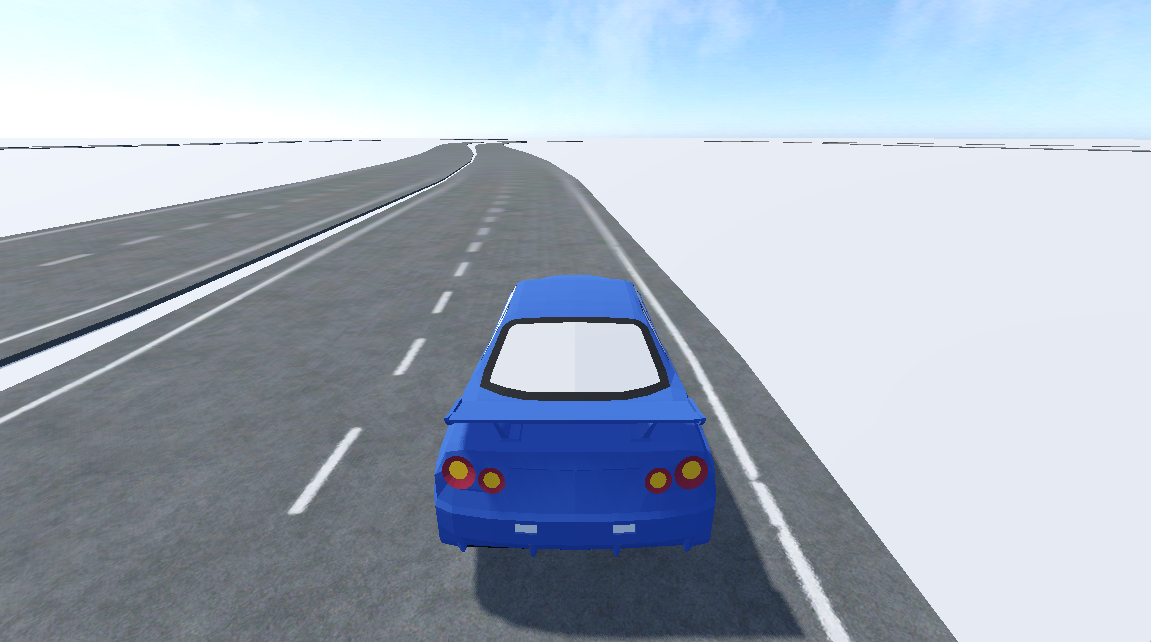}
        \caption{A car driving on the generated road}
    \end{subfigure}
    \caption{Extract the contour splines and apply modifications to them.}
\vspace{-0.4cm}   
\end{wrapfigure}


Godot is an open-source game engine \cite{godot} that offers exceptional flexibility for customized environment generation. Godot operates a UDP server on the back end, listening for incoming messages via a socket. Data transmission to Godot is managed by Python scripts. To transform a given image of a road into data interpretable by Godot, we use Python in conjunction with OpenCV to compute and extract the road's contour from the image. The extracted contour is then pixelated and saved to disk.

In a subsequent Python script, we iterate over each pixel in the image. If a pixel is white—indicating it is part of the road—we examine its neighboring pixels to calculate a value that determines the shape of the road segment to instantiate. This data is transmitted to Godot through a WebSocket. Upon receiving the data, Godot processes each value and instantiates the corresponding road shape within the scene. For determining which tile to use for the road, generation, we follow the instructions in \cite{road_gen}.

Figure \ref{fig:extract-contour} demonstrates that the vision detection system identifies the contour of the road. After fitting the contour to the road, we can reconstruct it in the simulator and position the target system, such as a car, for testing.

Using splines offers the advantage of modifying the road's shape under quantifiable constraints. For instance, the contour can be adjusted to ensure that the maximum deviation from the original road stays within a defined threshold. This method also enables the generation of various road variations.
\paragraph{Scenario Reconstruction Using Synthetic Roads}
We evaluate our methods using roads described by human input. The first image on the left illustrates the image generated from our handwritten data, while the second image demonstrates the reconstruction within the simulator.

\begin{wrapfigure}{r}{0.7\linewidth}
    \centering
    \begin{subfigure}{0.3\linewidth}
            \includegraphics[width=\linewidth]{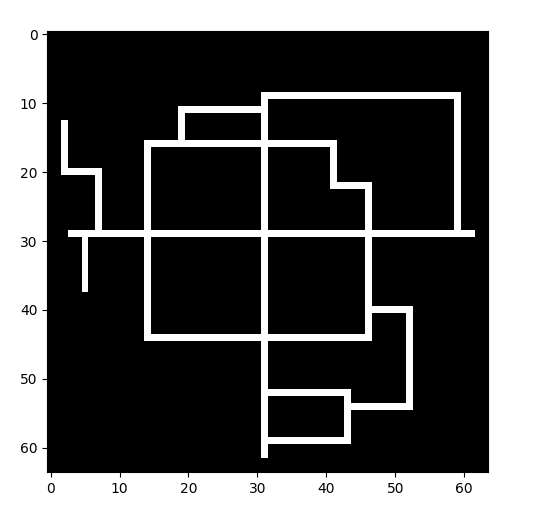}
    \caption{Basic pixel drawing}
    \end{subfigure}
    \begin{subfigure}{0.28\linewidth}
            \includegraphics[trim={1cm 0 1cm 1cm}, clip, 
width=\linewidth]{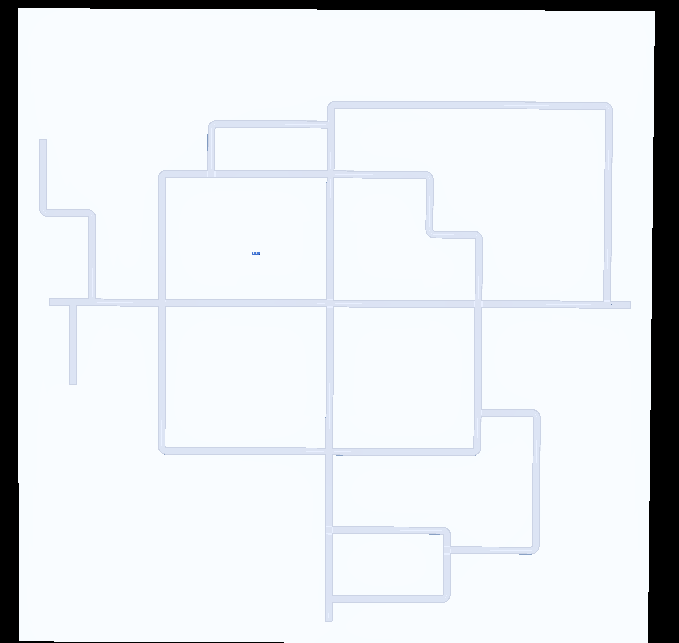}
\caption{Generated in simulator}
    \end{subfigure}
    \begin{subfigure}{0.3\linewidth}
            \includegraphics[ 
width=\linewidth]{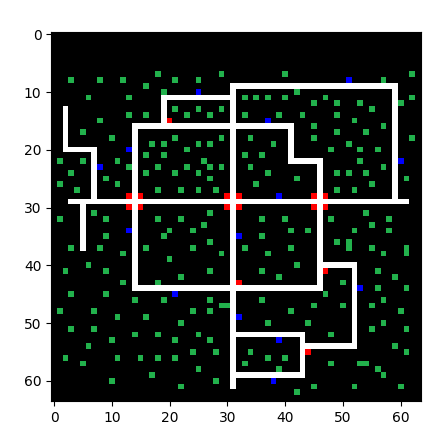}
\caption{Add new pixel colors}
    \end{subfigure}
\begin{subfigure}{0.4\linewidth}
        \includegraphics[ 
width=\linewidth]{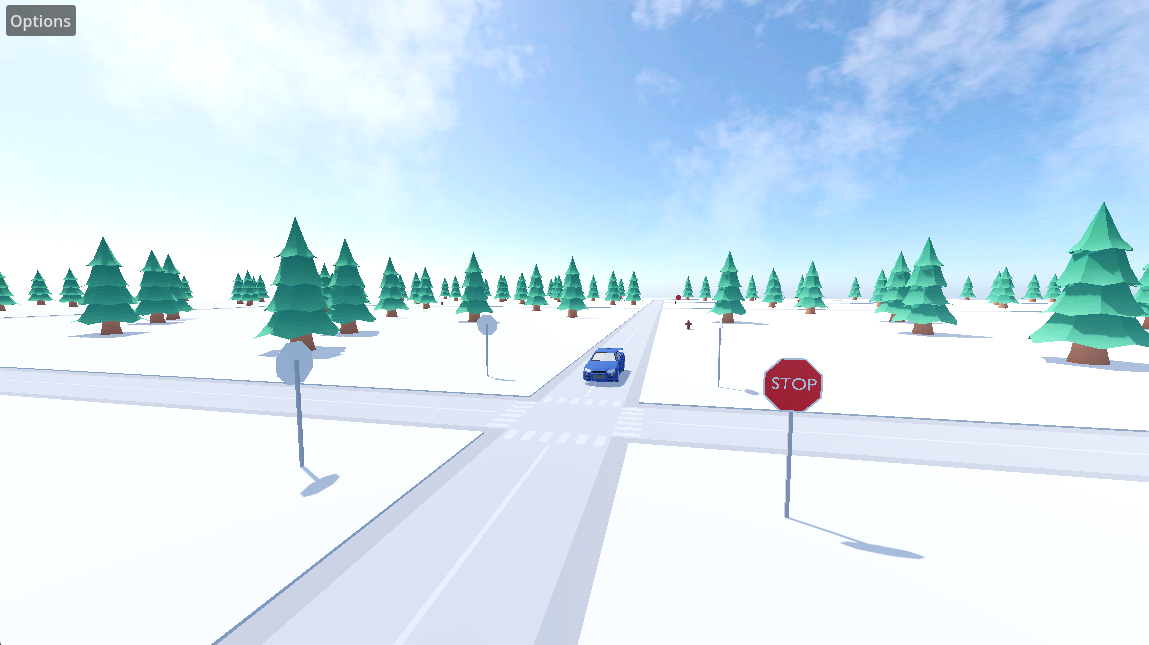}
\caption{New objects added to the scene in the simulator}
\end{subfigure}

    \caption{The first figure shows the pixelated version generated using OpenCV. The third figure illustrates how it is reconstructed in the Godot simulator.}
    \label{fig:godot-demo}

    \vspace{-1.2cm}
\end{wrapfigure}

Figure \ref{fig:godot-demo} shows that the computer vision algorithm accurately detected the shape of the road, and the simulator successfully generated the corresponding road.


\paragraph{Discussions} The primary limitation of the pipeline is the accuracy of contour detection in images. Various computer vision algorithms produce different levels of accuracy and each image may require a different threshold to accurately detect roads. Currently, we manually tune the threshold for each image, which may hinder scalability in future applications. Figure \ref{fig-contour-shadow} shows how shadows from trees can affect the contour, causing occlusions along the edge of the road. In the future, we will explore integrating more learning-based methods for road detection \cite{wang2021road} with our current pipeline. The ability to integrate depends on whether the representation of road locations can be interpreted as coordinates within the simulator.

\section{Related Work}\label{sec:related-work}
\paragraph{2D Image to 3D Photorealistic Scene Generation.} 2D image to 3D photorealistic scene generation has seen several advancements in recent years \cite{scenedreamer}, \cite{gancraft}. Traditionally, photogrammetry has been a primary method for rendering 3D scenes through stitching collections of photographic images and utilizing the camera’s spatial position to calculate measurements and locations of objects \cite{meshroom}, \cite{phototourism}.  While this method is effective and gives detailed results, it is limited to what the camera captured. Thus, it relies on high-quality input images, and any occlusions in captures result in incomplete models. Neural Radiance Fields (NeRFs) aimed to address the issues with photogrammetry. NeRFs use deep neural networks to synthesize additional scene information in real-time \cite{mipnerf360}. The drawback is that using NeRFs can be computationally expensive and slow. Specifically, altering the position of the camera forces the scene to be re-rendered each time. Most recently, Gaussian Splatting took the best of photogrammetry and NeRFs. This technique can model radiance fields using 3D Gaussians distributions \cite{gsplatting}. When given 3D data inputs, algorithms utilize neural networks to blend between data points with small Gaussians of varying shapes and densities to fill the gaps accurately. These methods can generate impressive and realistic 3D scenes, but there lacks a pipeline to easily convert the models into usable scenes from simulations. Especially when in need of large data sets of varying scenes, generating and converting 3D scenes can be time-consuming.
\paragraph{3D Scene Synthesis.} 3D scene synthesis has proven to be a highly promising method for generating environments for robot learning \cite{realisticscenesynthesis}, \cite{scenexproceduralcontrollablelargescale}, \cite{pipelinescenegen}. A large example is the popular household rearrangement problem. A space is defined – such as a room or a table-top – and an algorithm determines the position and orientation of randomly selected 3D models in that space \cite{mixeddiffusion3dindoor}, \cite{cluttergen}, \cite{sceneformerindoor}, \cite{controllablefurniturelayout}. However, objects are constrained to a fixed region. Furthermore, computational resources and complexity grows significantly as the quantity of objects increases. Thus, there is a lack in diversity when generating within a practical number of objects.
\paragraph{Game engines} Godot is a free and open-source game engine designed with a focus of allowing video game developers to easily create both 2D and 3D games \cite{godot}. Godot features a unique ecosystem that makes use of a hierarchy of nodes, where each node can be a class or represent entire scenes. Godot encourages to use the component-based system to facilitate development. Although video games are the primary focus of Godot, it is versatile enough to be used to develop non-game software applications as well. Unity \cite{Unity} and Unreal \cite{UnrealEngine} are two widely used game engines that serve as the foundation for well-known simulation environments such as CARLA and AirSim. While the level of customizability differs between the two engines, migrating our current pipeline to Unity or Unreal presents an intriguing future direction. Additionally, other engines excel in specific domains, such as modeling robotic behavior. For example, Mujoco \cite{todorov2012mujoco} is a physics engine with significant potential for integration with game engines to enhance simulations.\\

\section{Conclusion}
We introduce a scenario generation pipeline that constructs road models in a simulator directly from images. The pipeline allows for road modifications, and we propose a design to use formal methods to filter roads that meet specified constraints. This approach is valuable for reconstructing diverse road infrastructure layouts within simulation environments. The pipeline can potentially be integrated with existing scenario generation methods to generate various relationships between vehicles and different maps simultaneously.
 
\bibliographystyle{splncs04}
\bibliography{ref}

\end{document}

%% file: macros.tex
\newcommand{\traj}{\xi}
\newcommand{\reals}{\mathbb{R}}

\newcommand{\always}{\mathbf{G}}
\newcommand{\until}{\mathbf{U}}
\newcommand{\eventually}{\mathbf{F}}

\newcommand{\nnreals}{\mathbb R^{\geq 0}}